\documentclass[journal]{IEEEtran}
\usepackage{cite}
\usepackage{amssymb}
\usepackage{amsmath}
\usepackage{amsfonts}
\usepackage{color}
\usepackage{algorithmic}
\usepackage{algorithm}
\usepackage{amsthm}
\usepackage{graphicx}
\usepackage{epstopdf}
\usepackage{subfig}
\usepackage{booktabs}
\usepackage{lipsum}
\usepackage{cuted}
\usepackage{textcomp}
\usepackage{verbatim}
\usepackage{array}  
\usepackage{float}    
\usepackage{xcolor}

\usepackage{multicol}
\usepackage{amsmath}
\newtheorem{lemma}{Lemma}
\ifCLASSINFOpdf

\else

\fi
\begin{document}
	\setlength{\abovedisplayskip}{2pt}
	\setlength{\belowdisplayskip}{2pt}
        \title{Adaptive and Parallel Split Federated Learning in Vehicular Edge Computing}
        \author{Xianke Qiang, Zheng Chang,~\IEEEmembership{Senior~Member,~IEEE}, Yun Hu, Lei Liu,~\IEEEmembership{Senior~Member,~IEEE}, Timo H\"am\"al\"ainen,~\IEEEmembership{Senior~Member,~IEEE}
	   \thanks{X. Qiang and Z. Chang are with School of Computer Science and Engineering, University of Electronic Science and Technology of China, Chengdu 611731, China. Y. Hu is with National Demonstration Center for Experimental Electronic Information \& Communications Engineering Education, Xidian University, Xi'an 710071, China. Lei Liu is with the Guangzhou Institute of Technology, Xidian University, Guangzhou 510555, China. Z. Chang and T. H\"am\"al\"ainen are with Faculty of Information Technology, University of Jyv\"askyl\"a, P. O. Box 35, FIN-40014 Jyv\"askyl\"a, Finland. Corresponding author: Zheng Chang(email: zheng.chang@jyu.fi)
	}}
        \maketitle

\begin{abstract}
Vehicular edge intelligence (VEI) is a promising paradigm for enabling future intelligent transportation systems by accommodating artificial intelligence (AI) at the vehicular edge computing (VEC) system. Federated learning (FL) stands as one of the fundamental technologies facilitating collaborative model training locally and aggregation, while safeguarding the privacy of vehicle data in VEI. However, traditional FL faces challenges in adapting to vehicle heterogeneity, training large models on resource-constrained vehicles, and remaining susceptible to model weight privacy leakage. Meanwhile, split learning (SL) is proposed as a promising collaborative learning framework which can mitigate the risk of model wights leakage, and release the training workload on vehicles. SL sequentially trains a model between a vehicle and an edge cloud (EC) by dividing the entire model into a vehicle-side model and an EC-side model at a given cut layer. In this work, we combine the advantages of SL and FL to develop an Adaptive Split Federated Learning scheme for Vehicular Edge Computing (ASFV). The ASFV scheme adaptively splits the model and parallelizes the training process, taking into account mobile vehicle selection and resource allocation. Our extensive simulations, conducted on non-independent and identically distributed data, demonstrate that the proposed ASFV solution significantly reduces training latency compared to existing benchmarks, while adapting to network dynamics and vehicles' mobility.\par

\end{abstract}
	
	
\section{Introduction}
The Intelligent Transportation System (ITS) \cite{alam2016introduction} has become a promising way to improve transportation safety, traffic efficiency, and system autonomy \cite{zhang2023vehicle} as a result of the development of wireless communications and the Internet of Things (IoT). Researchers are increasingly focusing on vehicular edge intelligence (VEI), which is believed to help the development of ITS\cite{10133894}. Integrating AI technology into the VEC platform, which offers storage, computing, and network resources, enables the realization of the full potential of VEI\cite{zhou2019edge, wu2021digital}. To better utilize the large amounts of onboard data, conventional Machine learning (ML) has demonstrated its potential in diverse ITS applications, encompassing object detection, traffic sign classification, congestion prediction, and velocity/acceleration forecasting \cite{ye2018machine}. However, the conventional method of sending raw data to centralized servers for ML raises significant privacy concerns \cite{li2017multi} and requires large amounts of bandwidth for wireless communication. \par

Thus, a privacy-preserving distributed ML framework, Federated Learning (FL) \cite{konevcny2015federated}, is widely adopted in modern VEC systems to ensure higher automation levels en route, where moving vehicles need to make swift operational decisions \cite{pervej2023resource}. In the VEC system comprising connected and autonomous vehicles, FL locally trains the model and centrally aggregates the results. This approach leverages the data and onboard units of vehicles while allowing data processing and storage at Edge Cloud (EC) locations such as Roadside Units (RSUs) or Base Stations (BSs) \cite{10214588, yu2020mobility}.
    
Despite the advantages of FL in VEC systems, there remain several challenges on fully unlocking the its potential. One of the most significant difficulties is the high heterogeneity among the vehicles/clients involved in training \cite{yang2022flash}. Another primary concern of FL is how to protect user privacy since sensitive information can still be revealed from model parameters or gradients by a third-party entity or the server \cite{Shen2023}. Furthermore, with the development of AI, we have entered the era of large models, which means the data and algorithm are progressively growing in size and complexity. Training complete and large models on resource-constrained vehicles poses a significant challenge.\par

\begin{figure}[t]
        \centering
        \includegraphics[width=0.48\textwidth]{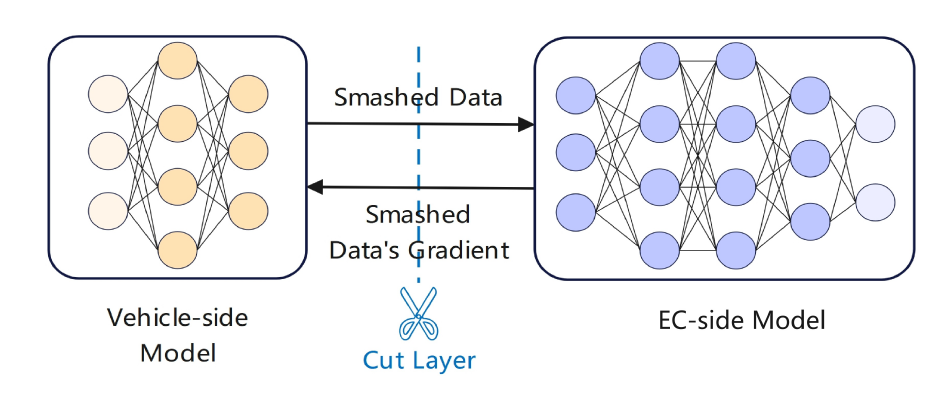} 
        \caption{SL splits the whole AI model into a vehicle-side model and a EC-side model at a cut layer (the third layer).}
        \label{fig:SL cutlayer} 
    \end{figure}

Split Learning (SL), as an emerging collaborative learning framework, facilitates the utilization of distributed vehicular data, reduces the risk of data leakage, and alleviates the training load on vehicles. SL was recently proposed in \cite{gupta2018distributed} and \cite{vepakomma2018split} by splitting the ML model (e.g., CNN) into several sub-models (e.g., a few layers of the entire CNN) with the cut layer and distributing them to different entities (e.g., the vehicle-side model at the vehicles or the EC-side model at the EC), which facilitates distributed learning via sharing the smashed data of the cut layer showing in Fig. \ref{fig:SL cutlayer}. The SL workflow mainly involves three main steps. Initially, the vehicle downloads the vehicle-side model and performs forward propagation to update its vehicle-side model, and transmits the processed data to the EC. Subsequently, the EC conducts backward propagation, updating the EC-side model and broadcasting the gradient associated with the cut layer back to vehicles. Every vehicle sequentially repeats the above process until all vehicles are down. The authors in \cite{9756883} have compared SL and FL with Transfer Learning (TL) and confirmed that the SL solution outperforms the other solutions in terms of accuracy, detection rates, the area under the curve, power consumption, packet delivery ratio, Quality of Experience (QoE) and delay analysis respectively, in the presence of malicious actions in the experienced ITS. \par 

    \begin{figure}[h]
        \centering
        \includegraphics[width=0.40\textwidth]{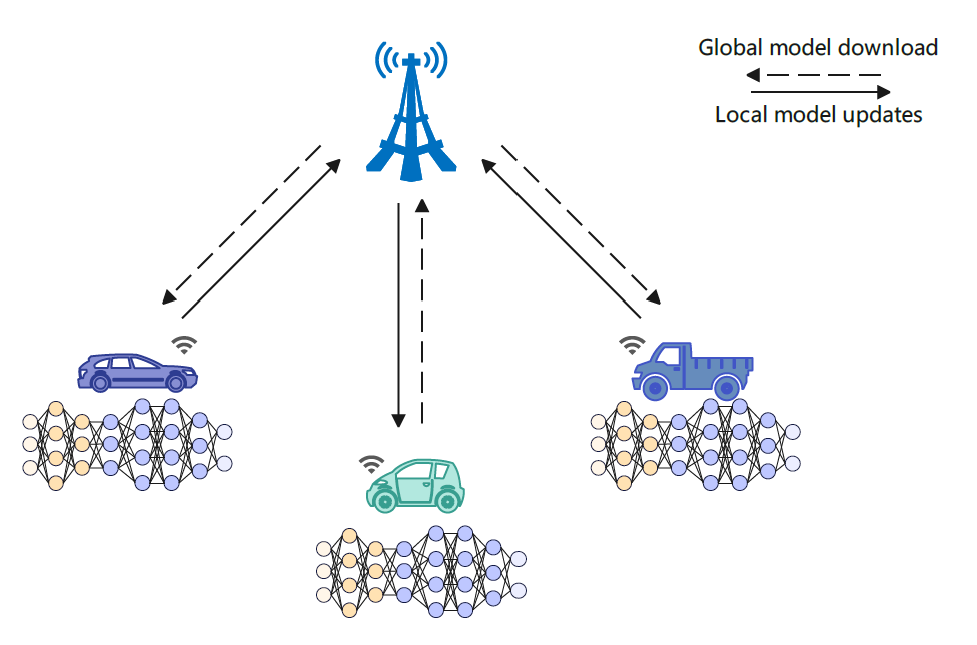}
        \caption{FL workflow}
        \label{fig:FL workflow}
        \hspace{1in}
        \includegraphics[width=0.40\textwidth]{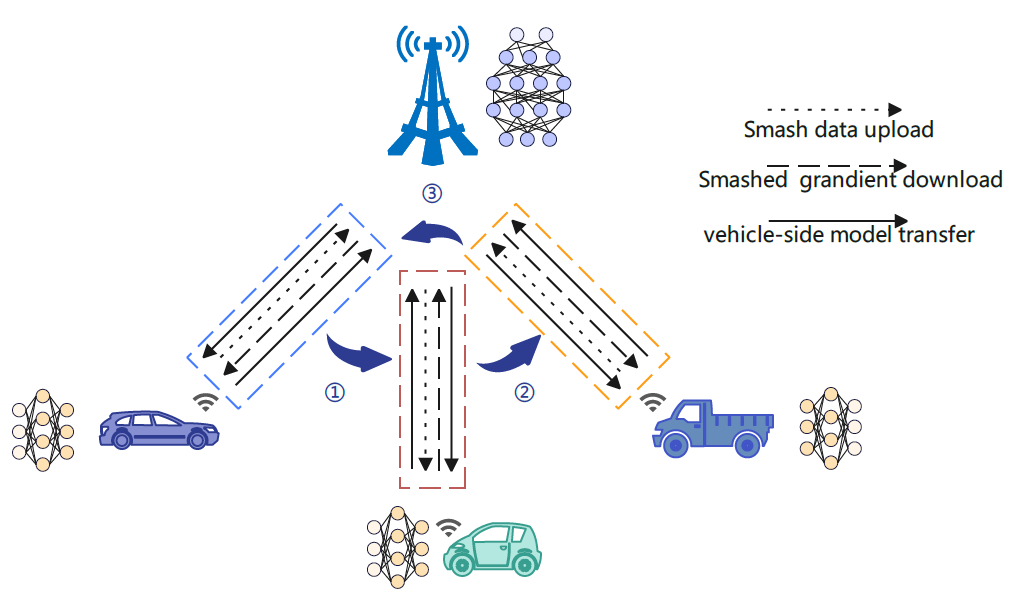}
        \caption{SL workflow}
        \label{fig:SL workflow}
    \end{figure}
    
However, utilizing the traditional sequential SL directly for VEC systems causing too many communication overload and time delays. A pioneering work called Split Federated Learning (SFL) \cite{thapa2022splitfed} combines the ideas of SL and FL to parallelize the training process. In this case, SFL not only reduces communication overhead and latency, but also reduces vehicle computing load, which make it more suitable for VEC systems. Different from FL, the research on SL and SFL is still in its infancy \cite{wu2023split}, especially in the area of  VEC systems. As most of the existing studies do not incorporate network dynamics, e.g., channel conditions, as well as vehicle computing capabilities, they don't consider how to obtain the optimal cut layer in real time. Besides, mobile vehicles traverse the range of the EC for varying durations, completing local training becomes challenging when these vehicles exit the current coverage area \cite{zhang2023vehicle} and the continuous movement of vehicles may hinder the timely uploading of local models to the EC, leading to potential delays in SL convergence and a reduction in model aggregation accuracy due to the dynamic nature of wireless channels. Moreover, the energy consumption for local computing is comparable to that for the wireless transmissions on mobile devices \cite{chen2022energy}. In this context, it is of great significance to choose the cut layer to meet the energy and time constraints, and also consider the resource allocation with mobile vehicles. \par

In this paper, we propose a SFL scheme, named \underline{A}daptive-\underline{S}plit \underline{F}ederated Learning for \underline{V}ehicular Edge Computing (ASFV), which parallels the vehicle-side model training while considering the vehicle mobility, unstable channel environment, and system time delay and energy consumption. To the authors’ knowledge, this article stands as the first work to fully illustrate the SFL in VEC system. The main contributions of this paper are summarized as follows:
    
 \begin{itemize}
 \item We propose a novel low-latency and low-energy ASFV by introducing an adaptive split federated training combining vehicle selection and resource allocation. Additionally, we conduct a thorough theoretical analysis of the training delay and energy consumption of the proposed ASFV. 
 
\item We propose a vehicle selection algorithm based on vehicle speed and EC communication range. And then a time delay minimization multi-objective function is formulated combining resource management considering vehicle heterogeneity, channel instability and model splitting strategy.

\item The formulated multi-objective problem is a mixed-interger non-linear programming and non-convex problem, which is NP-hard and very difficult to be directively solved. Consequently, we decompose the problem into three subproblems and iteratively solve the approximate optimal solution using BCD method. The three subproblems is online adaptive cut layer selection problem, transmission power assignment problem and wireless resource allocation problem, they solved by using KKT, SCA and Lagrange multiplier Method respectively.

\item We evaluate the performance of our proposed solution via extensive simulations using various open datasets to verify the effectiveness of our proposed scheme. Compared with existing schemes, our proposed method shows significant superiority in terms of time-energy efficiency and learning performance for ASFV over heterogeneous devices.   
\end{itemize}

\section{Related Work} 
\subsection{Federated Learning}
In 2016, Google proposed FL\cite{pmlr-v54-mcmahan17a}, and since then, it has become one of the most popular distributed learning methods. Numerous efforts have been dedicated to enhancing the performance of FL from various research perspectives. Some research papers \cite{liu2020client,luo2020hfel} explore the design of a multi-tier FL framework to effectively accommodate a substantial number of devices with a wider coverage range. To enhance the longevity of resource-constrained end devices, compression strategies such as weight quantization \cite{fu2021cpt,chen2022energy} and gradient quantization\cite{alistarh2017qsgd,chen2023service} are usually used to reduce computational complexity and communication overhead. To facilitate FL over dynamic wireless networks, several pioneering works have recently studied how to jointly optimize FL performance and cost efficiency, including communication efficiency and energy efficiency, in IoT systems. \cite{vu2020cell} jointly optimizes local accuracy, transmit power, data rate, and devices’ computing capacities to minimize FL training time. \cite{ren2020accelerating} jointly optimize local training batchsize and communication resource allocation to achieve fast training speed while maintaining learning accuracy. In \cite{hu2022communication}, they propose a communication-efficient federated learning framework with a partial model aggregation algorithm to utilize compression strategy and weighted vehicle selection, which can significantly reduce the size of uploaded
data and decrease the communication time.\par

\subsection{Split Learning}
SL is first proposed in 2018\cite{GUPTA20181}. SL has been widely used in the field of health care \cite{vepakomma2018split,poirot2019split}. 
From a communication perspective, SL performs slower than FL due to the relay-based training across multiple clients. Motivated by this, the SFL has been proposed in \cite{thapa2022splitfed}, which exploits the parallel model training mechanism in FL and the model splitting structure of SL. \cite{liu2022wireless} firstly propose a novel distributed learning architecture, a hybrid split and federated learning (HSFL) algorithm by reaping the parallel model training mechanism of FL and the model splitting structure of SL. Due to the dynamic communication environment and the development of 6G, the model training time and communication time can be compared, with the latter cut layer costing more training time but less communication time. How to choose cut layers has become particularly important. A pioneering work proposes an online learning algorithm to determine the optimal cut layer to minimize the training latency\cite{zhang2021learning}. \cite{wu2023split} design a novel SL scheme to reduce the training latency, named Cluster-based Parallel SL (CPSL) which conducts model training in a “first-parallel-then-sequential” manner. \par

\subsection{FL and SL for VEC System}

As one of the typical IoT systems, the FL performance and cost efficiency of the VEC system can be optimized with the above approaches. Many papers consider the FL-assisted VEC\cite{zhang2023vehicle,zeng2022federated,pervej2023resource,zhao2022participant}. \cite{zhang2023vehicle} propose a vehicle mobility and channel dynamic-aware FL (MADCA-FL) scheme to fit VEC systems, and formulate an MINLP problem to improve the learning performance under cost and resource budget constraints by jointly optimizing the computation and communication resources. The authors in \cite{zeng2022federated} propose a novel dynamic algorithm DFP to account for the varying vehicle participation and not independent and identically distributed (non-IID) data distribution among vehicles in the FL training process. \cite{pervej2023resource} presents a vehicular edge federated
learning framework with a joint study of the impact of the mobility of the clients with a practical 5G-NR-based RAT solution under strict delay, energy, computation resource, radio resource, and cost constraints. Newt\cite{zhao2022participant}, an enhanced federated learning approach including a new client selection utility explores the trade-off between accuracy performance in each round and system progress. \par

SL, as an emerging collaborative learning framework is still not fully investigated yet, especially in the vehicular network. In this work\cite{9756883}, a Split Learning-based IDS (SplitLearn) for ITS infrastructures has been proposed to address the potential security concerns and the proposed SplitLearn performed better than  Federated Learning (FedLearn) and Transfer Learning (TransLearn). They\cite{moon2023split} propose SplitFed learning with a mobility method to minimize the training time of the model, and a migration method for the ML model when the vehicle moves from the current serving VECs to the target VECs. 

However, currently in the SL-assisted and SFL-assisted VEC, there are still many problems that urgently need to be solved. Fistly, Although \cite{moon2023split} considered vehicle mobility and conducted model migration based on it, but only proposed rough ideas without conducting a detailed analysis. Secondly, the CPSL\cite{wu2023split} or HSFL\cite{liu2022wireless} architecture uses some devices for FL and some for SL, still not directly facing the problem of long SL serial delay. Thirdly, the offline selection\cite{wu2023split} of split layers is only determined by a split layer, if the subsequent vehicle movement or channel conditions change, cut layer offline selection is very likely not to be the current optimal cut layer. Different from the existing works, we focus on an adaptive and parallel SFL solution with an adaptive model  splitting for supporting a large number of vehicles. Furthermore, taking vehicle heterogeneity, network dynamics and vehicle mobility into account, we propose a resource management algorithm to optimize the performance of the proposed solution over wireless networks.
\begin{figure} 
    \centering 
    \includegraphics[scale=0.20]{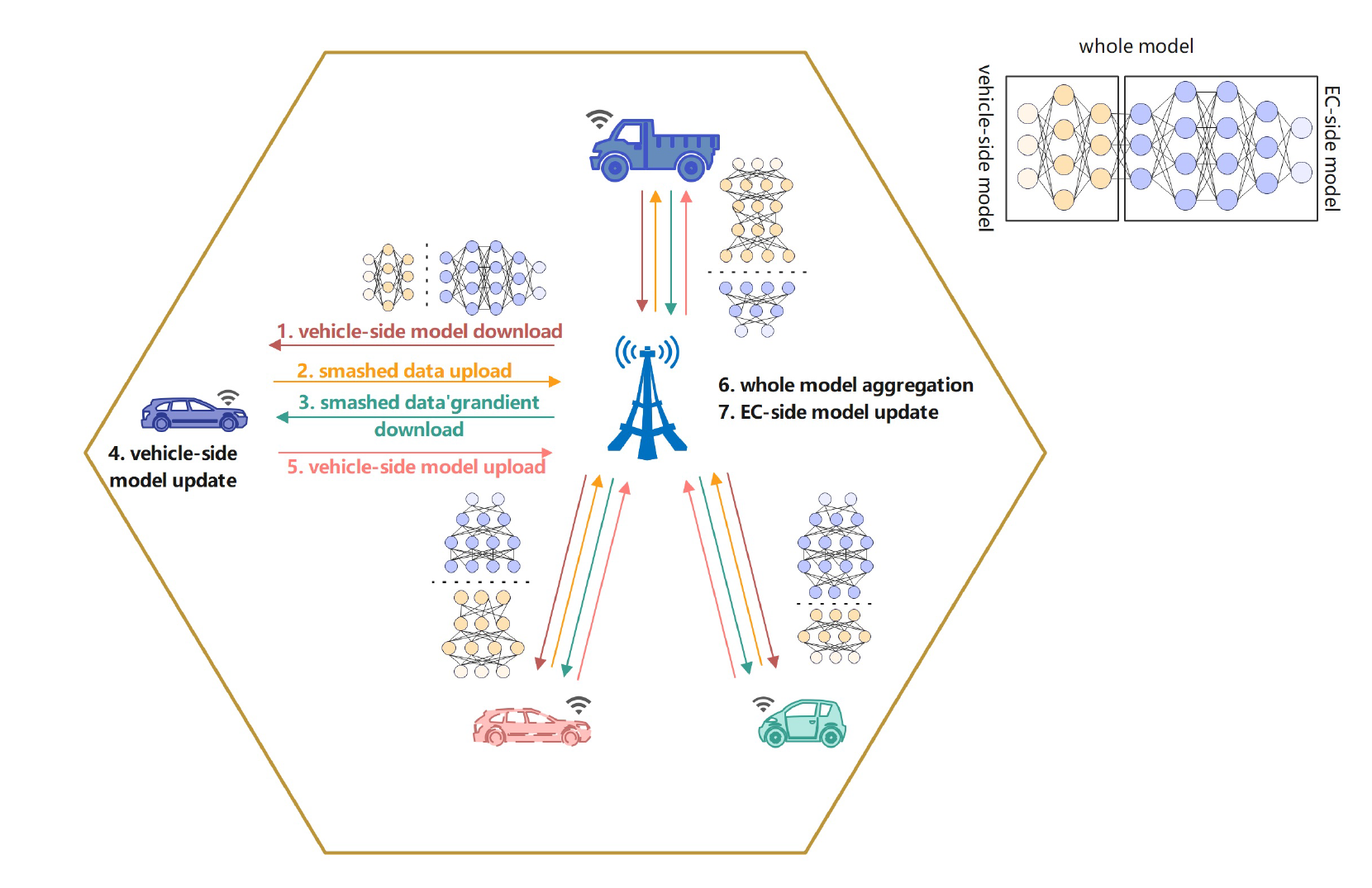}
    \caption{Split Federated Learning for Vehicle Network Workflow}
    \label{fig:work-flow}
\end{figure}
\section{System Model}
 As shown in Fig. \ref{fig:work-flow}, we introduce our new parallel scheme, called adaptive split federated learning for vehicular edge computing systems (ASFV). We consider a general VEC system that includes one EC, is deployed on RSUs or BSs and a set of vehicles $\mathcal{N}=\{1,2,\dots, N\}$. The set of available vehicles within the communication range of the EC at round $t$ is denoted by $\mathcal{N}_t$ which satisfies $\mathcal{N}_t \subset \mathcal{N}$. The data set of the vehicle $n$ is denoted as $\mathcal{D}_n = \{\mathcal{X}_n, \mathcal{Y}_n\}$, where $\mathcal{X}_n = \{x_{n}^{1}, x_{n}^2, ..., x_{n}^{ \left|\mathcal{D}_n\right|}\}$ is the training data, $\mathcal{Y}_n = \{y_{n}^1, y_{n}^2, ...,y_{n}^{ \left|\mathcal{D}_n\right|}\}$ represents the corresponding labels, and $\left|\mathcal{D}_n\right|$ is the number of training data samples of vehicle $n$. Firstly, different vehicle downloads different vehicle-side model $\omega_t^{V,\epsilon}$ according to different cut layer  $\{\epsilon, \epsilon \in \mathcal{E}\}$ set by EC, and execute forward propagation to upload the smashed data $A_t^{n,\epsilon}$ to the server. Secondly, the server-side model $\omega^{R,\epsilon}$ perform the forward and backward propagation with received smashed data, and then broadcasts the gradients of smashed data. Finally, the updated device-side model is upload to server for aggregation. 

\subsection{Computation Model}
 In this paper, we use $l(\omega,x_n^{i})$ denotes the loss function of each data sample $i$. For each dataset $\mathcal{D}_n$ of vehicle $n$, the local loss function of vehicle $n$ is 
                \begin{equation}
                    L_n(\omega) = \frac{1}{ \left|\mathcal{D}_n\right|} \sum_{i=1}^{ \left|\mathcal{D}_n\right|}l(\omega,x_{n}^i) 
                \end{equation}
        At the EC, the goal is to learn a model over the dataset distributed in $N$ vehicles, that is, the EC aims to obtain an optimal vector $\omega$ to minimize a loss function $L(\omega)$ by using the dataset distributed over all the vehicles. The objective of the considered learning task is to find the optimal model weight $\omega^*$ that minimize the global loss function $L(\omega)$: 
                \begin{equation}
                    \min_{\boldsymbol{\omega}} L(\boldsymbol{\omega})=\sum_{n=1}^N \rho_n L_n(\boldsymbol{\omega}),
                \end{equation} 
        where $\rho_n = \frac{ \left|\mathcal{D}_n\right|}{\sum_{n=1}^{N}  \left|\mathcal{D}_n\right|}$. \par The full model of vehicle $n$ in the $t$-th round $\omega_{t}^{n,\epsilon}$ includes two sub-models with $\epsilon$-th cut layer $\omega^{V,\epsilon}_{t}$ and $\omega^{R,\epsilon}_{t}$, it can be denoted by  
                \begin{equation}
                    \boldsymbol{\omega}_{t}^{n,\epsilon} = \{\omega^{V,\epsilon}_{t};\omega^{R,\epsilon}_{t}\},
                \end{equation} 
        and the global model update principle is as follows:
                \begin{equation}
                    \Delta \boldsymbol{\omega}_{t+1}^{n,\epsilon} = \boldsymbol{\omega}_{t+1}^{n,\epsilon} - \omega_{t},
                \end{equation}
                \begin{align}
               \boldsymbol{\omega}_{t+1} = \boldsymbol{\omega}_t - \sum_{n \in \mathcal{N}_t} p_n \Delta \boldsymbol{\omega}_{t+1}^{n,\epsilon},
            \end{align}
        where $p_n$ is the vehicle selection probability.
    
\subsection{Communication Model}
In this paper, we consider broadcasting for downlink transmission during the data transmission process. EC provides a total bandwidth $W$, The downlink transmission rate from EC to vehicle $n$ is the same and equal to
            \begin{equation}
                \label{R^{DL}}
                    R^{DL}= \underset{\forall n \in \mathcal{N}_t}{\min} \{W ln(1+\frac{h_r \phi_r d_{n,r}^{-\gamma}}{\sigma_0^2})\}, 
            \end{equation}
where $\sigma_0^2$ is the noise power, $h_r$ is channel gain of EC, $\phi_r$ is the transmission power of EC, $d^{-\gamma}_{n,r}$ represents the distance between vehicle $n$ and the EC, and $\gamma$ is the path loss exponent. 

We consider an orthogonal frequency-division multiple access (OFDMA) transmission protocol for uplink transmission during the data transmission process. We define $\beta_{n}$ as the bandwidth allocation ratio for vehicle $n$ such that EC's resulting allocated bandwidth is $\beta_{n} W$. Let $R_n^{UL}$ denote the achievable transmission rate of vehicle $n$ which is defined as
        \begin{equation}
            \label{R_n^{UL}}
            R_n^{UL}=\beta_{n} W ln(1+\frac{h_n \phi_n d_{n,r}^{-\gamma}}{\sigma_0^2}),
        \end{equation}
    where $\phi_n$ is the transmission power, and $h_n$ is the channel gain of vehicle $n$.
\subsection{Delay-Energy analysis}
    In this article, based on the above communication and computing models, we conduct a detailed delay energy consumption analysis on the model in the following.
        \subsubsection{Vehicle-side model distribution phase}
            The EC decides the cut layer $\epsilon_n^t$ according to the channel environment and splits the whole model in the $\epsilon_n^t$-th cut layer. Then, EC distributes the vehicle-side model to the selected vehicles respectively. The model distribute latency is given by
                \begin{align}
                    t_{d,n} &= \frac{s(\omega^{V,\epsilon_n^t})}{R^{DL}}.
                \end{align}
            
        \subsubsection{Vehicle-side model execution phase}
            The vehicle-side model execution refers to the vehicle-side model’s forward propagation process. Let $\gamma_v^F(\epsilon_n^t)$ denote the computation workload (in Flops) of vehicle-side model's forward propagation process for processing a data sample \cite{zhang2018shufflenet, zeng2021energy}, and $\kappa$ denotes the computing intensity, represents the number of Flops can be completed in one CPU cycle. To simplify the expression, we define $c_{v}^{F}(\epsilon_n^t) = \frac{\gamma_v^F(\epsilon_n^t)}{\kappa}$ to denotes the number of CPU cycles for vehicle $n$ to process one sample data forward propagation. Local training data size of vehicle $n$ is $\left| \mathcal{D}_n \right|$. The latency is given by 
                \begin{align}
                    t_{e,n} &= \frac{\left|\mathcal{D}_n\right|\gamma_v^F(\epsilon_n^t)}{f_n\kappa} =\frac{\left|\mathcal{D}_n\right| c_{v}^{F}}{f_n},\\
                    e_{e,n} & =\frac{\zeta}{2} \left|\mathcal{D}_n\right| c_{v}^{F}(\epsilon_n^t) f_n^2,
                \end{align}
            \noindent where $\zeta/2$ represents the effective capacitance coefficient of vehicle $n$'s computing chipset, $f_n$ denotes the central processing unit capability of vehicle $n$ \cite{wu2023split}.

        \subsubsection{Smashed data transmission phase}
            Each vehicle transmits the smashed data to the EC using the OFDMA method in which EC provides a total bandwidth $W$. The uplink transmission rate of vehicle $n$ is defined in (\ref{R^{DL}}). Let $s(A^{n,\epsilon_n^t})$ denote the smashed data size with respect to a data sample, also depending on cut layer $\epsilon_n^t$.  
                \begin{align}
                    \label{eq5}
                    t_{s,n} &= \frac{s(A^{n,\epsilon_n^t})}{R_n^{UL}},\\
                    e_{s,n} &= \phi_n t_{s,n}.
                \end{align}

        \subsubsection{EC-side model execution phase}
            The latency component includes two parts: (1) the first part is the time taken for performing the EC-side model’s forward propagation process, and (2) the second part is the time taken for performing the back propagation process of the EC-side model. Let $\gamma_r^F(\epsilon_n^t)$ and $\gamma_r^B(\epsilon_n^t)$ denote the computation workload of the EC-side model’s forward propagation and back propagation process for processing a data sample respectively, and the overall computation workload of $\left|\mathcal{D}_n\right|$ data samples is $\left|\mathcal{D}_n\right|\gamma_r^F(\epsilon_n^t)+\left|\mathcal{D}_n\right|\gamma_r^B(\epsilon_n^t)$. Taking the two parts into account, the overall latency is given by
                \begin{equation}
                    t_R = \frac{\left|\mathcal{D}_n\right| (\gamma_r^F(\epsilon_n^t)+\gamma_r^B(\epsilon_n^t))}{f_s \kappa},
                \end{equation}
            where $f_s$ represents the CPU frequency of the EC.
        \subsubsection{Smashed data’s gradient transmission phase}        
            Smashed data’s gradient $g(A^{\epsilon_n^t,n})$ is sent back to each vehicle using broadcasting. Since the time delay for a vehicle receiving smashed data's gradient is small compared to uploading local model parameters.
                \begin{align}
                    t_{g,n} &= \frac{s(g(A^{n,\epsilon_n^t}))}{R^{DL}},
                \end{align}
        \subsubsection{Vehicle-side model update phase}
            The vehicle-side model update refers to the back propagation process updating vehicle-side model parameters. Let $\gamma_v^B(\epsilon_n^t)$ represent the computation workload of the vehicle-side model’s back propagation process for a data sample. Let $c_{v}^{B}(\epsilon_n^t)$ be the number of CPU cycles for vehicle $n$ to process one sample data backward propagation. We have
                \begin{align}
                    t_{u,n} &= \frac{\left|\mathcal{D}_n\right| \gamma_v^B(\epsilon)}{f_n\kappa},\\
                    &=\frac{\left|\mathcal{D}_n\right| c_{v}^{B}(\epsilon)}{f_n}, \forall n \in \mathcal{N}_t,\\
                    e_{u,n}&=\frac{\zeta}{2} c_{v}^{B}(\epsilon) f_n^2.
                \end{align}
        \subsubsection{Vehicle-side model transmission phase}
            Let $s(\omega^{V,\epsilon_n^t})$ denote the data size (in bits) of the vehicle-side model.
                \begin{align}
                    t_{w,n} &= \frac{s(\omega^{V,\epsilon_n^t})}{R_n^{UL}}\\
                      e_{w,n}&=\phi_n t_{w,n} .  
                \end{align}
        \subsubsection{Overall time delay and energy consumption}
            To simplify the notations, we first introduce the following terms:
                \begin{align}
                    \overline{s_a(\epsilon)} &= s(A_t^{n,\epsilon}) + s(\omega^{V,\epsilon}),\nonumber\\
                    \overline{s_g(\epsilon)} &= s(g(A_t^{n,\epsilon})) + s(\omega^{V,\epsilon}), \nonumber\\
                    c_{v}(\epsilon) &= \frac{\gamma_v^F(\epsilon)}{\kappa} + \frac{\gamma_v^B(\epsilon)}{\kappa} = c_{v}^{F}(\epsilon) + c_{v}^{B}(\epsilon).\nonumber\\
                    c_{r}(\epsilon) &= \frac{\gamma_r^F(\epsilon)}{\kappa} + \frac{\gamma_r^B(\epsilon)}{\kappa} = c_{r}^{F}(\epsilon) + c_{r}^{B}(\epsilon).\nonumber
                \end{align}
            Thus, the overall time delay and energy consumption for vehicle $n$ is 
                \begin{align}
                    t_n &=\frac{\overline{s_g(\epsilon_n)}}{R^{DL}}+\frac{\left|\mathcal{D}_n\right| c_v(\epsilon_n)}{f_n}+\frac{\left|\mathcal{D}_n\right| c_r(\epsilon_n)}{f_r}+\frac{\overline{s_a(\epsilon_n)}}{R_n^{UL}},\\
                    e_n &={\frac{\zeta}{2}c_v(\epsilon_n)f_n^2\left|\mathcal{D}_n\right|}+\phi_n(\frac{\overline{s_a(\epsilon_n)}}{R_n^{UL}}).
                \end{align}
            The overall time delay of the ASFV parallel of the whole selected vehicles in one training round is
                \begin{align}
                    T = &\max_{n=1}^{\mathcal{N}_t}(\frac{\left|\mathcal{D}_n\right| c_v(\epsilon_n)}{f_n}+\frac{\overline{s_a(\epsilon_n)}}{R_n^{UL}}) \nonumber\\
                 +& \sum_{n=1}^{\mathcal{N}_t}(\frac{\overline{s_g(\epsilon_n)}}{R^{DL}}+\frac{\left|\mathcal{D}_n\right| c_r(\epsilon_n)}{f_r}). 
                \end{align}
            In the section IV, these decisions are optimized to minimize the training latency under energy constraints: vehicle heterogeneity, channel instability and cut layer selection.
  \subsection{Convergence Analysis}
        To analyze the convergence rate, we first make the assumptions as follows:\par
        \textbf{Assumption 1:}$L_1, \ldots, L_n$ are all $\ell$-smooth, i.e., for all $\boldsymbol{v}$ and $\boldsymbol{\omega}, L_n(\boldsymbol{v}) \leq L_n(\boldsymbol{\omega})+(\boldsymbol{v}-\boldsymbol{\omega})^T \nabla L_n(\boldsymbol{\omega})+\frac{\ell}{2}\|\boldsymbol{v}-\boldsymbol{\omega}\|_2^2$,\par
        \textbf{Assumption 2:} $L_1, \ldots, L_n$ are all $\mu$-strongly convex, i.e., for all $\boldsymbol{v}$ and $\boldsymbol{\omega}, L_n(\boldsymbol{v}) \geq L_n(\boldsymbol{\omega})+(\boldsymbol{v}-\boldsymbol{\omega})^T \nabla L_n(\boldsymbol{\omega})+$ $\frac{\mu}{2}\|\boldsymbol{v}-\boldsymbol{\omega}\|_2^2$\par
        \textbf{Assumption 3:} Let $\xi_t^n$ present the random sample dataset from the UE $u_n$. The variance of stochastic gradients in each UE is bounded: $\mathbb{E}\left\|\nabla L_n\left(\boldsymbol{\omega}_t^{n,\epsilon}, \xi_t^n\right)-\nabla L_n\left(\boldsymbol{\omega}_t^{n,\epsilon}\right)\right\|^2 \leq$ $\delta_n^2$, for $n=1, \ldots, N$\par
        \textbf{Assumption 4:} The expected squared norm of the stochastic gradients is uniformly bounded, i.e., $\mathbb{E}\left\|\boldsymbol{g}_n\left(\boldsymbol{\omega}_t^{n,\epsilon}, \xi_t^{n,\epsilon}\right)\right\|^2 \leq$ $G^2$, for $n=1, \ldots, N$\par
        \textbf{Assumption 5:} Assuming that $\mathcal{N}_t$ is a subset of $K$ vehicles uniformly sampled from $N$ vehicles without replacement. Assuming that the data is balanced and non-IID in the sense that $p_1=p_2=...=p_N=\frac{1}{N}$. The model aggregation performs as  $\boldsymbol{\omega}_{t+1}=\boldsymbol{\omega}_t$ - $\sum_{n \in \mathcal{N}_t} p_n \Delta \boldsymbol{\omega}_t^{n,\epsilon}.$\par
        \textbf{Convergence results:} Let Assumptions 1 to 5 hold, we assume $\varrho = \frac{2}{\mu}$ with $\iota=\frac{4 \ell}{\mu}$ and let $\nu=\frac{\ell}{\mu}$, the proposed ASFL algorithm with $K$ UEs selected for participation satisfies\cite{amiri2021convergence, liu2022wireless, li2019convergence}:
            \begin{equation}
                \mathbb{E}\left[L\left(\boldsymbol{\omega}_T\right)\right]-L^* \leq \frac{\nu}{\iota+T-1}\left(\frac{2 \Gamma}{\mu}+\frac{\mu \iota}{2} \mathbb{E}\left\|\boldsymbol{\omega}_1-\boldsymbol{\omega}^*\right\|^2\right),
            \end{equation}
        where $\Gamma =\sum_{n=1}^N p_n^2 \delta_n^2+6 \ell \gamma_v+8 G^2+$ $ (\frac{N}{K}-1)\frac{N}{N-1} G^2$, and the degree of non-IID can be represented $ \gamma_v = L^* - \sum_{n=1}^K p_n L_n^*$.\par
        Moreover, the convergence speed increases with the increasing number of selected vehicles on training.
        
\section{Vehicle Selection and Problem Formulation}
    In this section, we introduce the vehicle selection strategy fitting the realistic vehicular communication scenario and improving convergence speed. Then we introduce our SFL delay minimization problem considering vehicle heterogeneity, channel instability and model splitting strategy.

\subsection{Vehicle Selection Strategy}
Because of the restricted range of EC, some vehicles with high mobility may transit quickly and are unable to finish the training. Consequently, not all vehicles engage in the training during each round. Furthermore, convergence analysis reveals that accurately determining which vehicles are participating in the training is crucial. Therefore, we will select vehicles based on their mobility characteristics to ensure effective training outcomes. \par

The standing time is the driving time of the vehicle staying in the area of EC. It largely depends on the position and speed of connected vehicles. The long standing time in the coverage area promises that the training process can be completed and its results can be delivered.\par

We denote the velocity of vehicle $n$ as $v_n$, which is assumed to remain steady. The diameter of the EC coverage is $\mathcal{D}$, and the distance from vehicle $n$ to the entrance is  ${d}_{n}$. Given the maximum latency during one iteration $t_{max}$. We define the maximum duration \cite{yu2020mobility} for vehicle $n$ to complete the computation and communication tasks successfully as:
            \begin{equation}
            \label{eq11}
                \overline{t} = \min\{\frac{\mathcal{D}-{d}_{n}}{v_n},t_{max}\}
            \end{equation}
        Then we denote the indicator of being selected $\hat{\alpha}_{n}$ for vehicle $n$:
            \begin{equation}
                \label{eq16}
                \hat{\alpha}_{n} = 
                \begin{cases}
                1, & t_n \leq \overline{t}\\
                0, & \text { otherwise.}
                \end{cases}
            \end{equation}
        Next, we define the vehicle selection probability $p_{n}$ as:
            \begin{equation}
                \label{eq18}
                p_{n} = 
                \begin{cases}
                0, & \hat{a_n} = 0  \\
                \frac{1}{\sum_{n \in \mathcal{N}}\hat{a_n}}, & \hat{a_n} =1,
                \end{cases}
            \end{equation}
        where $p_n = 0$ means vehicle $n$ has no probability be selected to join in training. 
\subsection{Problem Formulation}
        \subsubsection{Vehicle heterogeneity} The computing and communication capabilities of vehicles are different, so the calculation frequency and transmission power of each vehicle in every training round are also different.
            \begin{align}
                 f_{min} &\leq f_n^t \leq f_{max}, \forall n \in \mathcal{N}_t, \nonumber \\
                \phi_{min}& \leq \phi_n^t \leq \phi_{max}, \forall n \in \mathcal{N}_t, \nonumber
            \end{align}
            represents different computing frequency $f_n$ and transmission power $\phi_n$ of vehicle $n$. 
        \subsubsection{Channel instability} 
            In every training epoch, we consider OFDMA for data transmission. Let $\beta_n$ represents the bandwidth allocation ratio of selected vehicle $n$ in $t$-th epoch. 
            \begin{align}
                 0 < \beta_n^t &\leq 1, \nonumber\\
                 \sum_{n \in \mathcal{N}_t} \beta_n^t &\leq 1. \nonumber
            \end{align}
            
        \subsubsection{Cut layer selection} 
            The deeper the cut layer, the larger vehicle-side model size and the smaller activations. The selection of the cut layer significantly influences both training and communication expenses. Here, we denote $\epsilon_n^t$ as the designated cut layer for vehicle $n$ in the $t$-th round.
            \begin{equation}
                \epsilon_n^t \in \mathcal{E},  \nonumber
            \end{equation}
            where $\mathcal{E} = \{2,...,8\}$ represents the number of available cut layer in our setting. \par
            The decision of selecting the cut layer is crucial, as it not only impacts the communication overhead due to the varying sizes of the vehicle-side model, the smashed data, and its gradient, which are dependent on the chosen cut layer, but also influences the distribution of computational workload between the vehicles and the edge cloud. Therefore, the selection of the cut layer plays a vital role in optimizing training latency \cite{wu2023split}.
        
        Based on the above decision variables, we consider minimizing training time delay in each round, and the problem can be found in the following. 
        \begin{align}
            \label{P}
    	\mathcal{P}: &\underset{\{\beta_n^t,f_n^t, \phi_n^t, \epsilon_n^t \}_{n \in \mathcal{N}_t}}{\min} T(\{p_n^t\}_{n\in\mathcal{N}},\{\beta_n^t,f_n^t,\phi_n^t,\epsilon_n^t\}_{n \in \mathcal{N}_t}) \nonumber \\
    	   &\text {s.t. }\text{C1: } \sum_{n=1}^{\mathcal{N}_t} \beta_n^t \leq 1 \nonumber\\
             &\text{C2: } 0 \leq \beta_n^t \leq 1, \nonumber\\
             &\text{C3: } e_n^t \leq \hat{E},\nonumber \\
             &\text{C4: } \epsilon_n^t \in \mathcal{E}, \nonumber\\
             &\text{C5: } f_{min} \leq f_n^t \leq f_{max}, \nonumber\\
             &\text{C6: } \phi_{min} \leq \phi_n^t \leq \phi_{max}, \nonumber\\
             &\text{C7: } \sum_{\mathcal{N}_t}p_n^t =1, 0 \leq p_n^t \leq 1. \nonumber
        \end{align}
        Constraints C1 and C2 ensure that each subchannel is solely allocated to one vehicle to avoid co-channel interference. C3 is the energy consumption upper bound of every vehicle per round. C4 shows the cut layer selection constraints,so the global model is partitioned into the vehicle-side model and the server-side model C5 is the computation frequency constraint of each vehicle and C6 shows the transmission power of vehicles cannot exceed its maximum. C7 shows the selection probability of each vehicles in each round, in the beginning of every round, the position and velocity of all vehicles $\mathcal{N}$ will be randomly reset and then selecting vehicles $\mathcal{N}_t$ according to the vehicle selection strategy in IV.A.
        
\section{Proposed Solution}
As we can see, $\mathcal{P}$ is a mixed-integer non-linear programming and obviously non-convex, which means $\mathcal{P}$ is NP-hard problem and it is very difficult to be directly solved. Therefore, we decompose the problem $\mathcal{P}$ into three subproblems and iteratively obtain the approximate optimal solution. 
    
\subsection{Online Adaptive Cut Layer Selection}
    Due to the cut layer $\epsilon_n^t \in \mathcal{E}$ is discrete and the variable space is small, so we can obtain the optimal cut layer for each training round by traversing method. Our objective function is to minimal the overall time cost of vehicle $n$.
    \begin{align}    
        \mathcal{SUBP} 1: \underset{\epsilon_n^t}{\min} &\{\frac{\left|\mathcal{D}_n\right| c_{v}(\epsilon_n^t)}{f_n} +\frac{\left|\mathcal{D}_n\right| c_r(\epsilon_n^t)}{f_r} \\& +\frac{\overline{s_g(\epsilon_n^t)}}{R^{DL}} + \frac{\overline{s_a(\epsilon_n^t)}}{R_n^{UL}}\}, \\
        &\text {s.t. }C3,C4.
    \end{align}
 
     The algorithm is presented in Algorithm \ref{alg1}.
    
     \begin{algorithm}[h]
            \caption{Adaptive cut layer selection} 
            \label{alg1}
            \begin{algorithmic}[1]
            \REQUIRE set of vehicle computation frequency $f_n$, bandwidth allocation ratio $\beta_n$ and transmission power $\phi_n$, max energy constraints $\hat{E}$.
            \FOR{every vehicle $n, n \in \mathcal{N}_t$}
            \FOR{every cut layer $e, e \in \mathcal{E}$}
            \STATE calculate the value of $\mathcal{SUB}1$ and record the cut layer corresponding to minimal value.
            \ENDFOR
            \ENDFOR
            \ENSURE	Set of optimal cut layer $\epsilon_n^t$ .
        \end{algorithmic}
    \end{algorithm}
   
\subsection{Optimal Transmission Power}

    To simplify the notations, we introduce the following terms:
        \begin{align}
            A &= \left|\mathcal{D}_n\right|c_{v}(\epsilon),\nonumber \\  
             C &=  \sum_{n \in \mathcal{N}_t} \frac{\overline{s_g(\epsilon)}}{R^{DL}} + \frac{\left|\mathcal{D}_n\right|c_r(\epsilon)}{f_r},\nonumber
        \end{align}
      
    Given the value of $\epsilon_n, f_n,\beta_n$, the transmission power subproblem can be converted as:
        \begin{align}
    	   \mathcal{SUBP}2:&\min \quad T(\{\phi_n^t\}_{n \in \mathcal{N}_t})\nonumber \\
            &\text {s.t. } C3,C6. \nonumber 
        \end{align}
        $\mathcal{SUBP}2$ is a min-max problem, so we give a upper bound time delay $\bar{T}$. Then we transfer the $\mathcal{SUBP}2$ into $\mathcal{SUBP}2'$ to minimize the upper bound $\bar{T}$ while $t_n<\bar{T},n \in \mathcal{N}_t$. 
        \begin{align}
            \mathcal{SUBP} 2':& \underset{\phi_n}{\min} \quad {\bar{T}} \label{subp2'}\\
            &\text {s.t. }C3,C6, \nonumber \\
            &C8:\frac{A}{f_n} + \frac{\overline{s_a(\epsilon_n)}}{\beta_n W ln(1+\frac{h_n \phi_n d_n^{-\gamma}}{\sigma_0^2})}  + C \leq \bar{T}.\nonumber
        \end{align}

    Obviously, we can see that C3 is non-convex, so we use SCA algorithm to obtain optimal transmission power        \cite{zhang2023vehicle}. We define $\phi_n^i$ as the uplink power of vehicle $n$ in $i$-th iteration and $e(\phi_n^i)$ as the energy consumption value $e_n$ of vehicle $n$ in the $i$-th iteration of SCA algorithm. To obtain the approximate upper bound, $e(\phi_n)$ can be approximated by its first-order Taylor expansion $\hat{e}(\phi_n^i,\phi_n)$ at point $\phi_n^i$, which is given by:
        \begin{equation}
        \label{subp2-1}
        \hat{e}(\phi_n^i,\phi_n) = e(\phi_n^i) + e'(\phi_n^i)(\phi_n-\phi_n^i),
        \end{equation}
    where $e'(\phi_n^i)$ is denoted as the first-order derivative of $e(\phi_n^i)$ at point $\phi_n^i$:
        \begin{align}
        \label{subp2-2}
            e'(\phi_n^i) &= -\frac{h_n d_n^{-\gamma}}{\beta_n W (\delta_0^2 + h_n \phi_n^i d_n^{-\gamma}) ln^2(1+\frac{h_n \phi_n^i d_n^{-\gamma}}{\sigma_0^2})} \nonumber \\
            &+\frac{\overline{s(\epsilon_n)}}{\beta_n W ln(a+\frac{h_n \phi_n^i d_n^{-\gamma}}{\sigma_0^2})}.
        \end{align}
    The problem is convex at each SCA iteration by changing $e(\phi_n)$ to $\hat{e}(\phi_n^i,\phi_n)$ in $\mathcal{SUBP}2'$. Then C3 can change into $ \hat{e} \leq \hat{E}$ to be convex. So we can obtain the optimal uplink power $\phi_n^{i,*}$ using Algorithm \ref{alg2}.

\subsection{Optimal Computation Frequency and Wireless Resource Allocation}
    To simplify the notations, we introduce the following terms:
        \begin{align}
            B &= \frac{\overline{s_a(\epsilon)}}{W ln(1+\frac{h_n \phi_n d_{n,r}^{-\gamma}}{\sigma_0^2})},\nonumber \\
             D &= \frac{\zeta}{2}\left| \mathcal{D}_n \right| c_{v}(\epsilon),\nonumber \\ 
            F &= \phi_n B.\nonumber 
        \end{align}
    Given the values of the cut layer $\epsilon$, transmission power $\phi_n$. The vehicle computing frequency and wireless resource allocation subproblem can be expressed as:
    \begin{align}
        \mathcal{SUBP} 3:& \underset{f_n,\beta_n}{\min} \quad \bar{T}\\
        \text {s.t. } &C1,C2,C3,C5,C8
    \end{align}
    For the optimization problem $\mathcal{SUBP}3$, we can show it is a convex optimization problem as stated in the following.
         \begin{algorithm}[t]
        \caption{Transmission Power Assignment using SCA Method} 
        \label{alg2}
        \begin{algorithmic}[1]
        \REQUIRE Set of $\epsilon_n$, $f_n$, $\beta_n$, max energy constraints $\hat{E}$, the initial uplink power $\phi_n^0$ of vehicle $n$, iteration round $i=0$, the accuracy requirement $\varepsilon$.
        \REPEAT
        \STATE calculate $\hat{e}(\phi_k^i,\phi_k)$ according to (\ref{subp2-1}),(\ref{subp2-2});
        \STATE solve $\mathcal{SUBP}2$ by substituting $\hat{e}(\phi_n^i)$ with $\hat{e}(\phi_n^i,\phi_n)$, and achieve the optimal solution $\phi_n^{i,*}$
        \STATE $\phi_n \rightarrow \phi_n^{i,*}$,$i \rightarrow i+1$
        \UNTIL{$\left\|\phi_n^i-\phi_n^{i-1}\right\| \leq \varepsilon$}
        \ENSURE	Optimal transmission power $\phi_n^*$.
        \end{algorithmic}
        \end{algorithm}    
      \begin{algorithm}[t]
        \caption{Resource Allocation using Lagrange Multiplier Method} 
        \label{alg3}
        \begin{algorithmic}[1]
        \REQUIRE Set $i = 0$, the initial Largrange multipliers set ($\mu_n^0$,$\tau^0$,$\sigma_n^0$), the step size $\eta_{\mu}$,$\eta_{\tau}$,$\eta_{\sigma}$,$\eta_{f}$;
        \REPEAT 
            \STATE Update the multiplier $\sigma_n^{i+1}$ as $\sigma_n^i+\eta_\sigma\frac{\partial L}{\partial \sigma}$,
            \STATE Update the multiplier $\mu_n^{i+1} = \frac{\sigma_n^{i+1}A}{2D {f_n^{i}}^3}$,
            \STATE Update the multiplier $\tau^{i+1} = \frac{\sigma_n^{i+1}B+\mu_n^{i+1}F}{{\beta_n^i}^2}$;
            \STATE Update the optimal $f_n^{i+1}=f_n^{i}-\eta_{f} \frac{\partial L}{\partial f_n}$;
            \STATE Update the optimal $\beta_n^{i+1}$ according to (\ref{eqkkt-7}) replaced with $\sigma_n^{i+1}$ and $f_n^{i+1}$;
            \STATE Update the optimal $\bar{T}^{i+1} = \max_{n \in \mathcal{N}_t} (\frac{A}{f_n^{i+1}}+\frac{B}{\beta_n^{i+1}}+C) $
        \UNTIL{Convergence}
        \end{algorithmic}
        \end{algorithm}

    \textit{Theorem 1:} The $\mathcal{SUBP}3$ is convex.\par
    \textit{Proof:} The subformulas of $\mathcal{SUBP}3$ consist of three parts: 1)$\frac{A}{f_n}$, 2)$\frac{B}{\beta_n}$ and 3)$Df_n^2+\frac{F}{\beta_n}$, each of which is intuitively convex in its domain and all constraints get affline such that problem $\mathcal{SUBP}3$ is convex. \par
    Since $\mathcal{SUBP}3$ is convex such that it can be solved by the Lagrange multiplier method. The partial Lagrange formula can be expressed as
        \begin{align}
            L_i =& \bar{T}+\tau(\sum_{n \in \mathcal{N}_t}\beta_n-1)+\sum_{n \in \mathcal{N}_t}\mu_n(D f_n^2+ \frac{F}{\beta_n}-\hat{E})\nonumber\\
            +&\sum_{n \in \mathcal{N}_t}\sigma_n(\frac{A}{f_n}+\frac{B}{\beta_n}+C-\bar{T}),\nonumber
        \end{align}
    where $\mu_n$, $\tau$ and $\sigma_n$ are the Lagrange multipliers related to constraints C1 and C3. Applying KKT conditions, we can derive the necessary and sufficient conditions in the following.
        \begin{align}
        \frac{\partial L_i}{\partial \beta_n} &=\tau - \frac{\sigma_nB+\mu_n F}{\beta_n^2} = 0,\label{eqkkt-1}\\
        \frac{\partial L_i}{\partial f_n} &=2\mu_n D f_n - \frac{A\sigma_n}{f_n^2} = 0,\label{eqkkt-2}\\
        \frac{\partial L_i}{\partial \bar{T}} &= 1- \sum_{n \in \mathcal{N}_t}\sigma_n = 0\\
        \tau(&\sum_{n \in \mathcal{N}_t}\beta_n -1)=0,\label{eqkkt-3}
    \end{align}
        \begin{align}
            \mu_n(&Df_n^2+\frac{F}{\beta_n}-\hat{E})=0,\\
            \sigma_n& (\frac{A}{f_n}+\frac{B}{\beta_n}+C-\bar{T})=0.
        \end{align}
    
    From (\ref{eqkkt-1}) and (\ref{eqkkt-2}), we can derive the relations below:
        \begin{align}
           \beta_n &= (\frac{\sigma_n B+\mu_n F}{\tau})^{\frac{1}{2}},\label{eqkkt-4}\\
            \tau&=\frac{\sigma_n B+\mu_n F}{\beta_n^2},\label{eqkkt-5}\\
            \mu_n &= \frac{A \sigma_n}{2D f_n^3},\label{eqkkt-6}
        \end{align}
    based on which, another relation expression can be obtained combining (\ref{eqkkt-3}) as follows.   
        \begin{equation}
            \beta_n = \frac{(\sigma_n B+\mu_n F)^{\frac{1}{2}}}{\sum_{n \in \mathcal{N}_t}(\sigma_n B+\mu_nF)^{\frac{1}{2}}}.
        \end{equation}
    Finally, replacing $\tau$ and $\mu_n$ with (\ref{eqkkt-5}) and (\ref{eqkkt-6}), the optimal bandwidth ratio $\beta_n^*$ can be easily solved out as following:
        \begin{equation}
        \beta_n^* = \frac{\sigma_n^{\frac{1}{2}}( B+\frac{A}{2D f_n^3} F)^{\frac{1}{2}}}{\sum_{n \in \mathcal{N}_t}\sigma_n^{\frac{1}{2}}( B+\frac{A}{2D f_n^3}F)^{\frac{1}{2}}},\label{eqkkt-7}
        \end{equation}
    Then we can find some closed-form solutions of variables as (\ref{eqkkt-5}), (\ref{eqkkt-6}) and (\ref{eqkkt-7}). The details are shown in Algorithm \ref{alg3}.
   \begin{algorithm}[t]
        \caption{Joint Optimal Algorithm using BCD method} 
        \label{alg4}
        \begin{algorithmic}[1]
        \REQUIRE Set $i=0$, $\epsilon_1,\epsilon_2,\epsilon_3>0$, $\mathcal{E}$. 
        \REPEAT
        \STATE Applying vehicle selection strategy to choose participant vehicles in this round.
        \STATE Choose the optimal cut layer \textbf{$\epsilon^{i}$} from $\mathcal{SUB}1$ at given $\phi^{i-1},\beta^{i-1},f^{i-1}$ using Algorithm \ref{alg1};
        \STATE Compute the optimal transmission power $\phi^{i}$ from $\mathcal{SUB}2$ at given $\epsilon^{i},\beta^{i-1},f^{i-1}$ by applying Algorithm \ref{alg2};
        \STATE Compute the optimal computation frequency $f^{i}$, $\beta^i$ and $\bar{T}^i$ at given $\epsilon^{i},\phi^{i}$
        \UNTIL{$\Vert \phi^i-\phi^{i-1}\Vert<\epsilon_1$,  $\Vert f^i-f^{i-1} \Vert<\epsilon_2$, $\Vert \beta^{i}-\beta^{i-1}\Vert<\epsilon_3$} by applying the Lagrange multiplier method;
        \STATE $i =  i + 1$;
        \ENSURE $\epsilon_n^{i},\phi_n^{i},\beta_n^{i},f_n^{i}$. 
        \end{algorithmic}
    \end{algorithm}
        \begin{figure}[t]
            \centering
            \includegraphics[width=0.48\textwidth]{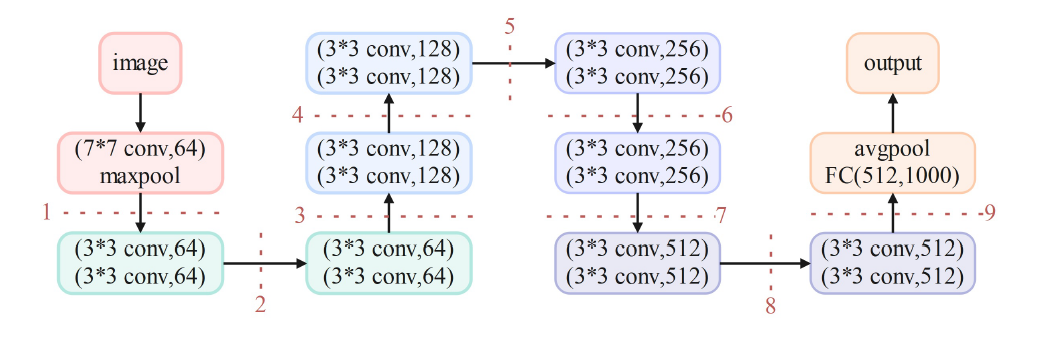} 
            \caption{ResNet18 Model Structure}
        \label{fig:resnet18} 
        \end{figure}
        \begin{table}[t]
                \centering
                \caption{ResNet18 Model Parameters}  
                \label{table1}  
                \begin{tabular}{cccc}
                \hline Cut Layer& $\gamma_v^F$/GFLOPs & $\gamma_r^F$/GFLOPs  \\
                \hline 
                0 & 0.00 & 14.89 \\ 
                1 & 0.99 & 13.90 \\
                2 & 2.89 & 12.00 \\
                3 & 4.79 & 10.10 \\
                4 & 6.27 & 8.62 \\
                5 & 8.16 & 6.72 \\
                6 & 9.64 & 5.25 \\
                7 & 11.53 & 3.36 \\
                8 & 13.00 & 1.89\\
                9 & 14.89 & 0.00 \\
                \hline
                \end{tabular}
            \end{table}
        \begin{figure}[t]
            \centering
            \includegraphics[width=0.35\textwidth]{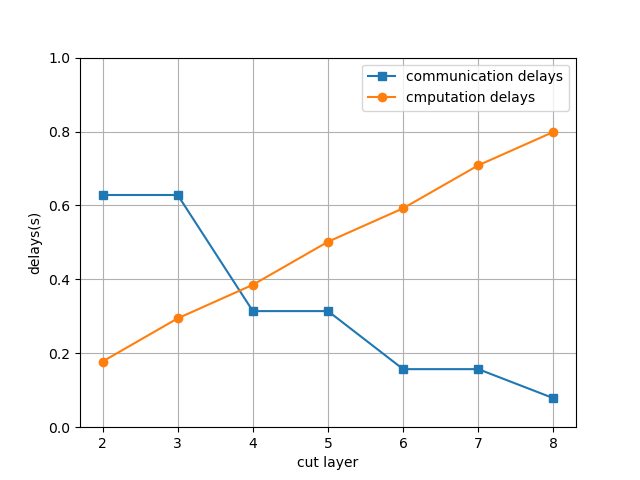} 
            \caption{Time delay with different cut layer}
            \label{fig:time delay with different cutlayer} 
        \end{figure}

    \subsection{Joint Algorithm}
    Although there is not a closed-form solution for the optimal power and wireless resource allocation, the block coordinate descent (BCD) approach can be used to find the optimal solutions. In Algorithm \ref{alg4}, $i$ initially defined as $i=0$. Firstly, EC applying the vehicle selection strategy to choose vehicles. According to select vehicles, then to solve $\mathcal{P}$, the optimal cut layer $\epsilon_n^i$ is obtained by fixing $\phi_n^{i-1},\beta_n^{i-1},f_n^{i-1}$ in the $i$-th iteration. The optimal transmission power $\phi_n^{i}$ is calculated by given $\epsilon_n^i,\beta_n^{i-1},f_n^{i-1}$, the value of $f_n^i$ is optimized with $\epsilon_n^i,\phi_n^{i},\beta_n^{i-1}$. Then $\beta_n^{i}$ can be directly calculated based on $f_n^i$. The loops end until the differences meet the threshold requirment $\epsilon_1,\epsilon_2,\epsilon_3$. \par
    The computation complexity of Algorithm \ref{alg4} mainly composed with the three subproblem\cite{bejaoui2020qos}. The complexity of vehicle selection strategy is $O(N)$, where $N$ is the number of selected vehicles. The complexity of $\mathcal{SUBP}1$ is $\mathcal{O}(EK)$, where $E$ is the number of cut layer and $K$ is the number of selected vehicles. According to \cite{boyd2004convex}, the $\mathcal{SUBP}2$ use the SCA method, and the computational complexity is $\mathcal{O}(I_{SCA}M^3))$, where $M$ is the number of variables. The complexity of $\mathcal{SUB}3$ using Lagrange Multiplier is $\mathcal{O}(M^{3.5}\log(\frac{1}{\epsilon}))$  according to\cite{bomze2010interior}. Therefore, the overall computation complexity of the overall algorithm is $\mathcal{O}(I_{BCD}*(N+EK+I_{SCA}M^3+M^{3.5}\log(\frac{1}{\epsilon})))$,where $I_{BCD}$ is the number of iterations of BCD algorithm.

    \begin{figure*}[t]
        \centering
        \subfloat[Mnist non-iid]{\includegraphics[width=0.33\textwidth]{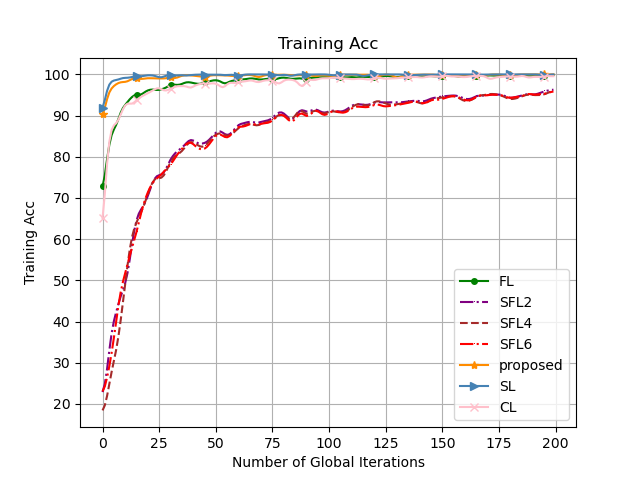}}
        \hfill
        \subfloat[Fashion-Mnist non-iid]{\includegraphics[width=0.33\textwidth]{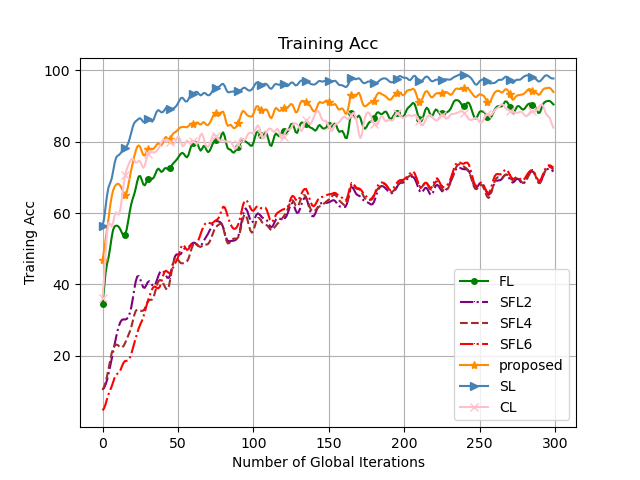}}
        \hfill
        \subfloat[CIFA10 non-iid]{\includegraphics[width=0.33\textwidth]{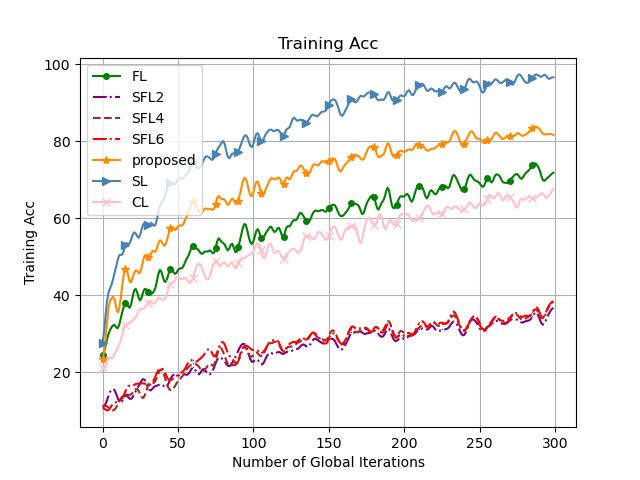}}
        \caption{Training Results of different wireless distributed learning algorithm with different datasets.}
        \label{fig:TrainingAcc}
    \end{figure*}
    \begin{figure*}[t]
        \centering
        \subfloat[Mnist non-iid]{\includegraphics[width=0.33\textwidth]{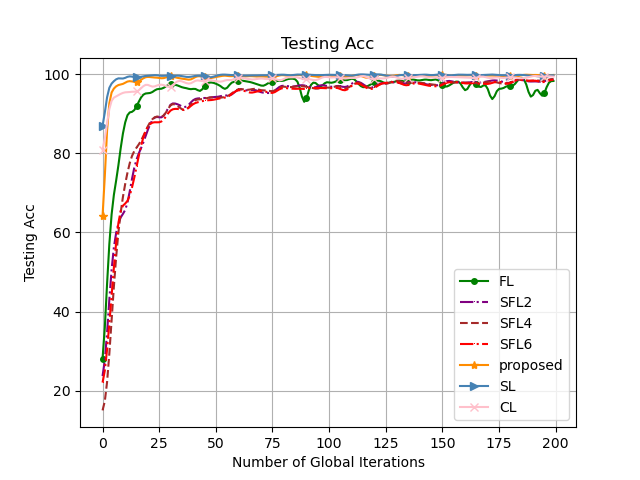}}
        \hfill
        \subfloat[Fashion-Mnist non-iid]{\includegraphics[width=0.33\textwidth]{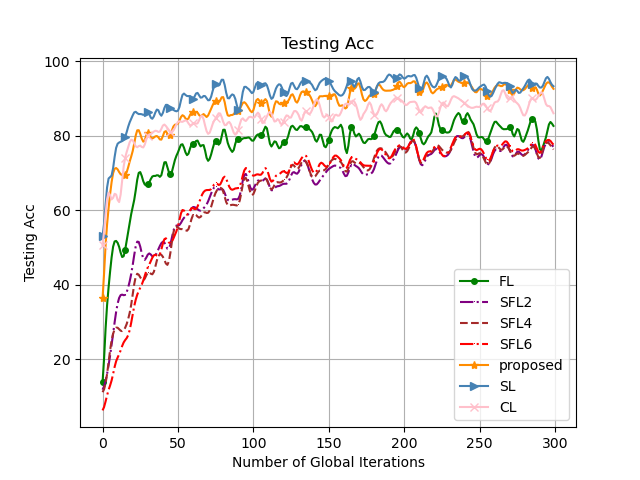}}
        \hfill
        \subfloat[CIFA10 non-iid]{\includegraphics[width=0.33\textwidth]{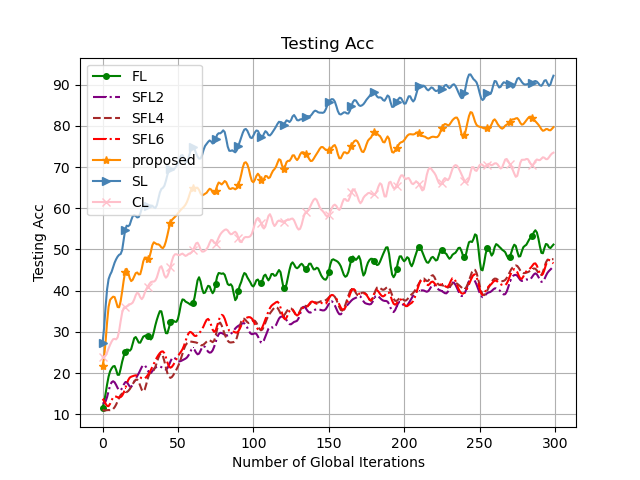}}
        \caption{Testing Results of different wireless distributed learning algorithm with different datasets.}
        \label{fig:TestingAcc}
    \end{figure*}
    
\section{Performance Evaluation}
 We evaluate the performance of the proposed ASFV scheme and resource management algorithm through comprehensive simulations.
\subsection{Simulation Setup}
 \subsubsection{Datasets}
     Our simulation leverages three distinct image classification datasets: (1) the MNIST dataset \cite{lecun1998gradient}, comprising images of handwritten digits from "0" to "9," each associated with a corresponding label; (2) the Fashion-MNIST dataset \cite{xiao2017fashion}, consisting of images representing various clothing items like "Shirt" and "Trouser," also labeled accordingly; and (3) the CIFAR-10 dataset \cite{krizhevsky2009learning}, containing colored images categorized into classes such as "Airplane" and "Automobile." Each dataset comprises a training set with 50,000 samples for model training and a test set with 10,000 samples for performance evaluation. Notably, data distribution at vehicles is non-IID, which widely exists in practical systems. To capture the heterogeneity among mobile vehicles in these datasets, we impose a constraint where each vehicle retains only three out of the ten possible labels, with sample sizes varying according to a power law as described in \cite{li2020federated}. We employ different learning rates for each dataset: 1e-5 for MNIST, 2e-6 for Fashion-MNIST, and 5e-6 for CIFAR-10. The local batch size is fixed at 64, and we conduct local training epochs $\tau$, set to 5.
    \subsubsection{ResNet model split strategy}
        We choose the ResNet18 as the global model. Resnet18 is compose with residual blocks\cite{he2016deep}. So in this simulation, we split the whole model by residual blocks. As shown in Fig. \ref{fig:resnet18}, we present the model split strategy. There are a total of 9 cut layers here, but we focus solely on the selection of cut layers 2 through 8 in our simulations. 

        The specific forward propagation workload for both vehicle-side and EC-side processing at different cut layers is detailed in Table \ref{table1}. According to \cite{katharopoulos2018not}, the backward propagation requires about twice the amount of time as the forward propagation since it needs to compute full gradients. Consequently, we define the backward propagation workloads as twice the forward propagation workloads.
        
      \begin{figure}[h]
        \centering
        \includegraphics[width=0.35\textwidth]{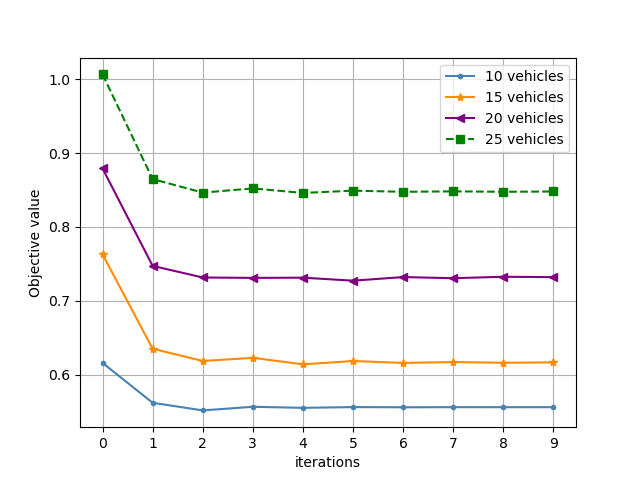} 
        \caption{The convergence of objective in $\mathcal{P}$.}
        \label{fig:cnvergence of the objective value} 
    \end{figure}
    \subsubsection{Communication and computation setting}
        The CPU frequency of vehicles is calculate by solving $\mathcal{SUBP}3$ and the value range from $10$ GHz to $20$ GHz. The CPU frequency of EC server is $50$ GHz. The number of transmission power $\phi_n$ for vehicle $n$ to process one sample data is randomly setting from $20$ dBm to $30$ dBm. The transmission power of EC is $40$ dBm. The noise power $\sigma_0^2$ is -100 dBm. 
        The number of vehicles in the coverage follows a Possion Distribution. We randomly selected vehicles with different channel conditions within a 500 meter adius of the EC, and calculated the average computation and communication time of these vehicles under different cut layers. In Fig. \ref{fig:time delay with different cutlayer}, we can see with the number of cut layer increasing, the average computing time for every vehicles is increasing. While the communication time for vehicles is decrease in $4,6,8$-th cut layer as the number of cut layers increases, we can observe that the smashed data size will decrease accordingly. Under the same channel conditions, as the number of split layers increases, the communication time will decrease.
               \begin{figure*}[t]
    \centering
    \subfloat[communication delays]{
            \includegraphics[width=0.30\textwidth]{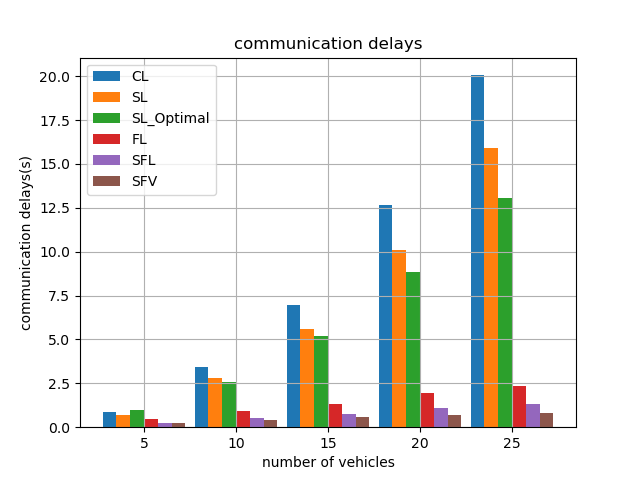}
            \label{commdelays}
        }
    \hfill
    \subfloat[computation delays]{
            \includegraphics[width=0.30\textwidth]{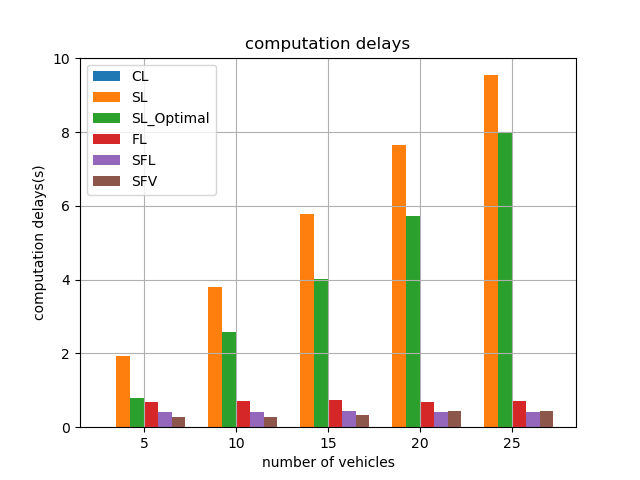}
            \label{compdelays}
    }
    \hfill
    \subfloat[overall delays]{
            \includegraphics[width=0.30\textwidth]{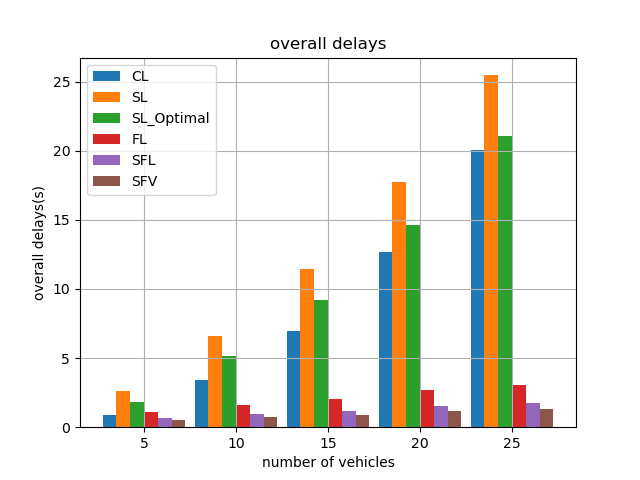}
            \label{overalldelays}
        }
    \caption{Communication and computation delays}
    \label{fig:delays}
    \end{figure*}
    \begin{figure*}[t]
        \centering
        \subfloat[communication energy]{
                \includegraphics[width=0.30\textwidth]{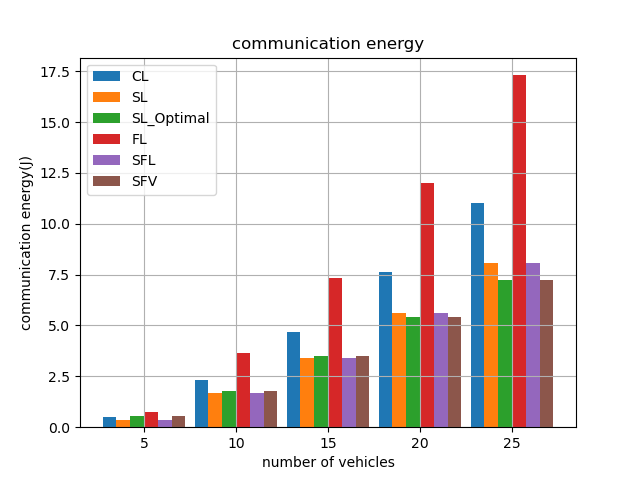}
                \label{commenergy}
            }
        \hfill
        \subfloat[computation energy]{
                \includegraphics[width=0.30\textwidth]{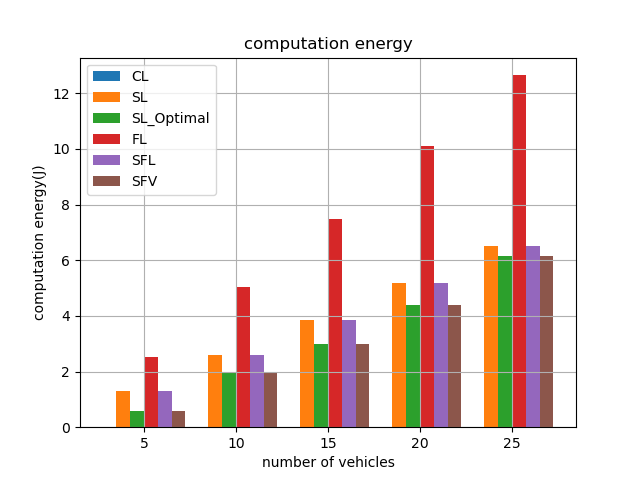}
                \label{compenergy}
        }
        \hfill
        \subfloat[overall energy]{
                \includegraphics[width=0.30\textwidth]{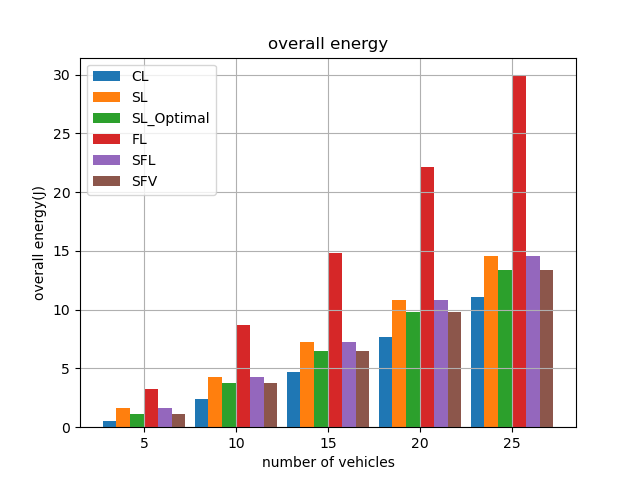}
                \label{overallenergy}
        }
        \caption{Communication and computation energy}
        \label{fig:energy}
    \end{figure*}

    \subsection{Performance Evaluation of the Proposed Scheme}
        Four baselines that are considered for the comparison with ASFL are given as following:
        \begin{itemize}
            \item CL: All vehicles transmit the raw images to the EC for training and aggregation.  
            \item FL \cite{konevcny2015federated}: Each vehicle trains its local model using raw images, then uploads the models to the EC for aggregation and updating.
            \item SL \cite{DBLP:journals/corr/abs-1912-12115}: The entire model is split into two parts—one for vehicles (vehicle-side model) and one for the EC (EC-side model). Each vehicle communicates sequentially with the EC to jointly train the entire model.
            \item SFL \cite{thapa2022splitfed}:  Similar to SL, the entire model is split into two parts for vehicles and the EC. Vehicles train the vehicle-side model in parallel and aggregate parameters to obtain a new global model at the EC.
        \end{itemize}
        \par
       To compare the performance of FL and SL among the above mentioned schemes over the MNIST, Fashion-MNIST and CIFAR10 datasets, we plot the values of the training accuracy and testing accuracy in Fig. \ref{fig:TrainingAcc} and Fig. \ref{fig:TestingAcc} respectively. Several notable observations emerge from our experiments, conducted under the same mobile vehicle selection strategy as proposed. Notably, our scheme showcases accuracy levels comparable to SL, despite our approach processing data in parallel rather than sequentially. This distinction highlights the outstanding performance of our approach. Specifically, our proposed scheme demonstrates significantly better performance than traditional SFL. Here, SFL2, SFL4, and SFL6 represent traditional SFL with the 2nd, 4th, and 6th cut layers, respectively. Additionally, SL, which splits the model into two parts with the 2nd cut layer, serves as a reference point for comparison.

        In the Fig. \ref{fig:TestingAcc}, our proposed scheme exhibits accuracy closest to SL, alongside significantly faster convergence speed compared to traditional FL and SFL methods. Notably, Parallel ASFV achieves notably higher accuracy than other parallel FL and SFL approaches, albeit slightly lower than sequential SL.

            In Fig. \ref{fig:cnvergence of the objective value}, we show the values of $T(\beta,f,\phi,\epsilon)$ under varying number of vehicles. It can be observed that under varying number of vehicles, as the number of iteration rounds grows, the objective value continues to decrease until it converges to a given level.

            Fig. \ref{fig:delays} illustrates communication, computation, and total time delays across five scenarios with varying selected vehicles ($N = 5, 10, 15, 20, 25$) in each round. The CL algorithm involves raw image processing updates. In contrast, the SL algorithm encompasses vehicle-side model distribution, vehicle-side model training, smashed data uploading, and vehicle-side model uploading. Notably, standard SL lacks resource optimization, whereas SL$\_$optimal incorporates optimal resource management. Meanwhile, SFL2, SFL4, SFL6, and our proposed ASFV focus on vehicle-side model training and uploading. However, ASFV distinguishes itself by integrating optimal resource allocation and cut layer selection, elements absent in FL and SFL algorithms.
            Fig. \ref{fig:delays}\subref{commdelays} shows the communication delays. CL incurs the longest time delays because all vehicles must upload their raw images to the EC in each scenario. SL follows with the second-highest time delays across all scenarios due to its sequential communication process among vehicles. Fig. \ref{fig:delays}\subref{compdelays} illustrates the computation delays. SL brings the longest training time to complete one epoch as vehicles are serially training. The CL algorithm don't cost computation time because the vehicles only transmit the raw images to EC for training. We can observe that SL$\_$optimal spend a few more seconds on computation compared with SL showing our resource allocation algorithm performs well. Fig. \ref{fig:delays}\subref{overalldelays} depicts the overall delays with different algorithm under varying number of vehicles. The SL method costs the longest delays to achieve one training epoch as vehicles having serial communication. The CL has the second highest time delays over all the scenarios because in which all the vehicles have to upload their raw images to the EC in each scenario. The training time of FL and SFL increases with increasing number of vehicles because the bandwidth allocated to each vehicles is decreasing. The SL experience increasing training time performance since the total bandwidth is fixed and the time delays mainly depends on the communication latency and number of vehicles. The ASFV increase slightly with the number of vehicles because the whole model is split according to the channel environment. \par
    
            Fig. \ref{fig:energy} presents the total energy consumption over selected vehicles $N = 5, 10, 15, 20, 25$ in five scenarios when training the ResNet18 model in one round. In these five scenarios, the energy consumption takes into account the sum of all vehicle computing and communication costs whether it is serial training or parallel training whether it is serial design or parallel design.\par
            Fig. \ref{fig:energy}\subref{commenergy} shows that the communication energy with varying number of vehicles. When there is 5 or 10 vehicles join in the training, we can observe that the communication energy consumption is comparable in all five scenarios. As more vehicles participate, the energy consumption for the system to complete one round of training becomes higher and higher. Fig. \ref{fig:energy}\subref{compenergy} illustrates the computation energy with incresing number of vehicles. The energy consumption of FL is the highest because the FL need to upload the whole model to EC. The energy consumption of SFL is higer than that of SFV as the SFL choose the stable cut layer while SFV choose the time-efficient optimal cut layer every epoch. And the communication energy consumption of SL$\_$Optimal is much less than that of SL without resource allocation optimal. Fig. \ref{fig:energy}\subref{overallenergy} describes the overall energy consumption with varying number of vehicles. The overall energy consumption of CL is the smallest as the CL only upload the raw data to the EC. The energy consumption of FL is the higher than SL, but the SL is much energy efficient. Our proposed SFV is not only time saving but also energy efficient.\par

\section{Conclusion}
    In this paper, we propose a novel low-latency and low-energy split federated learning scheme, namely adaptive split federated learning for vehicular edge computing (ASFV) by introducing adaptive split model and parallel training procedure combining the vehicle selection and resource allocation. By applying our ASFV algorithm in vehicular networks, our results demonstrated it achieved higher learning accuracy than FL and SFL, and less communication time delays and energy consumption. Additionally, we conduct a thorough theoretical analysis of the training latency and energy consumption of ASFV. Our simulation results demonstrated it achieved higher learning accuracy than FL and near SL accuracy, and less communication overhead than FL and SL under independent and identically non-IID data.

\appendix
In this section, we examine the ASFV scheme under the conditions of partial UEs participation on non-IID data. We define $\boldsymbol{g}_t = \sum_{n=1}^Np_n\boldsymbol{g}_t^n(\omega_t^{n,\epsilon},\xi_t^n)$ and $\overline{\boldsymbol{g}}_t = \sum_{n=1}^Np_n\overline{\boldsymbol{g}}_t^n(\omega_t^{n,\epsilon},\xi_t^n)$, thus, $\mathbb{E}\boldsymbol{g}=\overline{\boldsymbol{g}}_t$.\par
\begin{figure}[h]
    \begin{align}
    \label{appendix-3}
    \Vert \omega_{t+1} - \omega^* \Vert^2 
    &= \Vert \omega_{t+1} - v_{t+1} + v_{t+1} - \omega^* \Vert^2 \nonumber\\ &=\underbrace{\Vert \omega_{t+1} - v_{t+1} \Vert^2}_{A_1}  + \underbrace{\Vert v_{t+1} - \omega^* \Vert^2}_{A_2} \nonumber\\ &+\underbrace{2 \langle \omega_{t+1} - v_{t+1}, v_{t+1} - \omega^*\rangle}_{A_3} 
    \end{align}
\end{figure}
From (\ref{appendix-1}), we bound the average of the terms $A_1,A_2$ and $A_3$. They are explained in three Lemmas where the proof of each is included.
    \begin{lemma}
    To bound $A_1$, we have the equation (\ref{appendix-1})
    \begin{equation}
    \label{lemma-2}
        \mathbb{E} \left\|\boldsymbol{\omega}_{t+1}-\boldsymbol{v}_{t+1}\right\|^2 \leq (\frac{N}{K}-1)\frac{N}{N-1}\eta_t^2G^2 
    \end{equation}
\end{lemma}
    \begin{lemma}
    To bound the $A_2$ by bounding the three terms $B_1,B_2$ and $B_3$, so we have the equation (\ref{appendix-2},\ref{appendix-B1},\ref{appendix-B2},\ref{appendix-C1}).
    \begin{align}
        B_1:&\left\|\omega_t-\omega^{\star}-\eta_t \overline{\mathbf{g}}_t\right\|^2 \\& \leq\left(1-\mu \eta_t\right)\left\|\omega_t-\omega^{\star}\right\|^2 +2 \sum_{k=1}^N p_k\left\|\omega_t-\omega_t^k\right\|^2+6 \eta_t^2 \ell \gamma_v \nonumber \\
        B_2:&\mathbb{E}\left\|\boldsymbol{g}_t-\overline{\boldsymbol{g}}_t\right\|^2  \leq \sum_{n=1}^N p_n^2 \delta_n^2 \\
        B_3:&2\mathbb{E}\left[\left\langle\boldsymbol{\omega}_t-\omega^*-\eta_t \overline{\boldsymbol{g}}_t, \eta_t \boldsymbol{g}_t-\eta_t \overline{\boldsymbol{g}}_t\right\rangle\right] = 0
    \end{align}
    \end{lemma}
    \begin{lemma}
    To bound $A_3$, let $\mathbb{E}_{\mathcal{N}_t}$ denote expectation over the vehicle selection randomness at t-th round $t$. We have
        \begin{equation}
            \label{lemma-1}
            \mathbb{E}_{\mathcal{N}_t}[\omega_{t+1}] =  v_{t+1} \nonumber
        \end{equation}
    from which it follows that 
        \begin{equation}
            \mathbb{E}_{\mathcal{N}_t}[<\omega_{t+1}-v_{t+1},v_{t+1}-\omega^{\star}>] = 0 
        \end{equation}
    So $A_3$ is bound as
        \begin{equation}
            2 \langle \omega_{t+1} - v_{t+1}, v_{t+1} - \omega^*\rangle = 0
        \end{equation}
    \end{lemma}
    
    \begin{figure*}[htbp]
        \begin{align}
        \label{appendix-1}
        A_1: \mathbb{E}\left\|\boldsymbol{\omega}_{t+1}-\boldsymbol{v}_{t+1}\right\|^2 
        &=\mathbb{E}\left\|\frac{1}{K} \sum_{n \in \mathcal{N}_{t+1}} \boldsymbol{\omega}_{t+1}^{n,\epsilon}-\boldsymbol{v}_{t+1}\right\|^2 =\frac{1}{K^2} \mathbb{E}\left\|\sum_{n=1}^N \mathbb{I}\left\{n \in \mathcal{N}_{t+1}\right\}\left(\boldsymbol{\omega}_{t+1}^{n,\epsilon}-\boldsymbol{v}_{t+1}\right)\right\|^2 \nonumber\\ 
        & =\frac{1}{K^2} \mathbb{E}_{\mathcal{N}_t}\left[\sum_{n \in N} \mathbb{P}\left(n \in \mathcal{N}_{t+1}\right)\left\|\boldsymbol{\omega}_{t+1}^{n,\epsilon}-\boldsymbol{v}_{t+1}\right\|^2+\sum_{i \neq j} \mathbb{P}\left(i, j \in \mathcal{N}_{t+1}\right)\left\langle\boldsymbol{\omega}_{t+1}^{i,\epsilon}-\boldsymbol{v}_{t+1}, \boldsymbol{v}_{t+1}^{i,\epsilon}-\boldsymbol{v}_{t+1}\right\rangle\right] \nonumber \\ 
        & =\frac{1}{K N} \sum_{n=1}^N \mathbb{E}\left\|\boldsymbol{\omega}_{t+1}^{n,\epsilon}-\boldsymbol{v}_{t+1}\right\|^2+\sum_{i \neq j} \frac{K-1}{K N(N-1)} \mathbb{E}\left\langle\boldsymbol{\omega}_{t+1}^{i,\epsilon}-\boldsymbol{v}_{t+1}, \boldsymbol{v}_{t+1}^{j,\epsilon}-\boldsymbol{v}_{t+1}\right\rangle \nonumber\\ 
        & =\frac{N}{K(N-1)}(1-\frac{K}{N}) \sum_{n=1}^N \mathbb{E}\left\|\boldsymbol{\omega}_{t+1}^{n,\epsilon}-\boldsymbol{v}_{t+1}\right\|^2 \nonumber\\ 
        & =\frac{N}{K(N-1)}(1-\frac{K}{N}) \mathbb{E}\left[\frac{1}{N} \sum_{n=1}^N\left\|\left(\boldsymbol{\omega}_{t+1}^{n,\epsilon}-\boldsymbol{\omega}_{t_0}\right)-\left(\boldsymbol{v}_{t+1}-\boldsymbol{\omega}_{t_0}\right)\right\|^2\right] \nonumber\\ 
        & =\frac{N}{K(N-1)}(1-\frac{K}{N}) \mathbb{E}\left[\frac{1}{N} \sum_{n=1}^N\left\|\boldsymbol{\omega}_{t+1}^{n,\epsilon}-\boldsymbol{\omega}_{t_0}\right\|^2\right] \nonumber\\ 
        & \leq (\frac{N}{K}-1)\frac{N}{N-1}\eta_t^2G^2 
        \end{align}
         {\noindent} \rule[-10pt]{18cm}{0.05em}
    \end{figure*}
    \begin{figure*}[htb]
    \begin{align}
    \label{appendix-2}
    A_2:\mathbb{E}\left\|\boldsymbol{v}_{t+1}-\boldsymbol{\omega}^*\right\|^2 =\mathbb{E}&\left\|\boldsymbol{\omega}_t-\eta_t \boldsymbol{g}_t-\boldsymbol{\omega}^*\right\|^2=\mathbb{E}\left\|\boldsymbol{\omega}_t-\eta_t \boldsymbol{g}_t-\boldsymbol{\omega}^*-\eta_t \overline{\boldsymbol{g}}_t+\eta_t \overline{\boldsymbol{g}}_t\right\|^2 \nonumber\\
        & =\mathbb{E} \underbrace{\left\|\boldsymbol{\omega}_t-\boldsymbol{\omega}^*-\eta_t \overline{\boldsymbol{g}}_t\right\|^2}_{B_1}+\eta_t^2 \underbrace{\mathbb{E} \left\|\boldsymbol{g}_t-\overline{\boldsymbol{g}}_t\right\|^2}_{B_2}+\underbrace{2 \mathbb{E}\left[\left\langle\boldsymbol{\omega}_t-\omega^*-\eta_t \overline{\boldsymbol{g}}_t, \eta_t \boldsymbol{g}_t-\eta_t \overline{\boldsymbol{g}}_t\right\rangle\right]}_{B_3}  \nonumber\\
        &\leq\left(1-\mu \eta_t\right)\left\|\omega_t^{n,\epsilon}-\omega^{\star}\right\|^2+2 \underbrace{\mathbb{E}\sum_{n=1}^N p_n\left\|\omega_t-\omega_t^{n,\epsilon}\right\|^2}_{C_1}+6 \eta_t^2 \ell \gamma_v+\eta_t^2 \sum_{n=1}^N p_n^2 \delta_n^2 
    \end{align}
     {\noindent} \rule[-10pt]{18cm}{0.05em}
    \end{figure*}
    \begin{figure}
        \begin{align}
        B_1:&\left\|\omega_t-\omega^{\star}-\eta_t \overline{\mathbf{g}}_t\right\|^2 \nonumber \\
        =& \left\|\omega_t-\omega^{\star}\right\|^2 -2 \eta_t\left\langle\omega_t-\omega^{\star}, \overline{\mathbf{g}}_t\right\rangle+\eta_t^2\left\|\overline{\mathbf{g}}_t\right\|^2 \nonumber\\
        &\leq\left(1-\mu \eta_t\right)\left\|\omega_t-\omega^{\star}\right\|^2+2 \sum_{k=1}^N p_k\left\|\omega_t-\omega_t^k\right\|^2+6 \eta_t^2 \ell \gamma_v 
        \label{appendix-B1}
        \end{align}
         {\noindent} 
    \end{figure}
    \begin{figure}
            \begin{align}
            B_2:&\mathbb{E}\left\|\boldsymbol{g}_t-\overline{\boldsymbol{g}}_t\right\|^2 \nonumber\\
            & =\mathbb{E}\left\|\sum_{n=1}^N p_n\left(\nabla L_n\left(\boldsymbol{\omega}_t^{n,\epsilon}, \xi_t^{n,\epsilon}\right)-\nabla L_n\left(\boldsymbol{\omega}_t^{n,\epsilon}\right)\right)\right\|^2\nonumber \\
            &=\sum_{n=1}^N p_n^2 \mathbb{E}\left\|\left(\nabla L_n\left(\boldsymbol{\omega}_t^{n,\epsilon}, \xi_t^{n,\epsilon}\right)-\nabla L_n\left(\boldsymbol{\omega}_t^{n,\epsilon}\right)\right)\right\|^2\nonumber \\
            & \leq \sum_{n=1}^N p_n^2 \delta_n^2 
            \label{appendix-B2}
        \end{align}
          {\noindent}
    \end{figure}
    \begin{figure}
        \begin{align}
        C_1&:\mathbb{E} \sum_{n=1}^N p_n\left\|\boldsymbol{\omega}_t-\boldsymbol{\omega}_t^{n,\epsilon}\right\|^2 \nonumber\\
        &= \mathbb{E} \sum_{n=1}^N p_n\left\|\left(\boldsymbol{\omega}_t^{n,\epsilon}-\boldsymbol{\omega}_{t_0}\right)-\left(\boldsymbol{\omega}_t-\boldsymbol{\omega}_{t_0}\right)\right\|^2 \nonumber\\
        &\leq \mathbb{E} \sum_{n=1}^N p_n\left\|\left(\boldsymbol{\omega}_t^{n,\epsilon}-\boldsymbol{\omega}_{t_0}\right)\right\|^2 \nonumber\\
        &\leq \sum_{\tau=t_0}^{t-1} \sum_{n=1}^N p_n \mathbb{E}\left\|\eta_\tau \boldsymbol{g}_\tau^{n,\epsilon}\right\|^2 \nonumber\\
        &\leq 4 \eta_t^2 G^2 
        \label{appendix-C1}
        \end{align}
        {\noindent}
    \end{figure}

According to \cite{liu2022wireless,li2019convergence}, we can get $\mathbb{E}\left\| \omega_{t+1}-\omega^*\right\| \leq (1-\eta_t^2\mu)\mathbb{E}\left\|\omega_t-\omega^*\right\|$. And we use the similar steps as in \cite{liu2022wireless,li2019convergence} and get the upper.

\bibliographystyle{IEEEtran}
\bibliography{main.bib}
	
\end{document}